\documentclass[10pt,twocolumn,letterpaper]{article}

\usepackage{iccv}
\usepackage{times}
\usepackage{epsfig}
\usepackage{graphicx}
\usepackage{amsmath}
\usepackage{amssymb}
\usepackage{xcolor}
\usepackage{mathtools}
\usepackage{svg}
\usepackage{booktabs}  % for nicer tables
\usepackage{siunitx}  % for nice number+unit formatting and automatic rounding in tables
\usepackage{nicefrac}

\DeclareMathOperator{\EX}{\mathbb{E}}% expected value
\DeclareMathOperator{\Var}{\text{Var}}% expected valuevariance

\usepackage[pagebackref=true,breaklinks=true,letterpaper=true,colorlinks,bookmarks=false]{hyperref}

\iccvfinalcopy % *** Uncomment this line for the final submission

\def\iccvPaperID{AtIKsJerIl} % *** Enter the ICCV Paper ID here

\ificcvfinal\fi

\newcommand{\todo}[1]{%
    \ifx\empty#1\empty
        \textcolor{red}{TODO}%
    \else
        \textcolor{red}{TODO: #1}%
    \fi
}

\begin{document}

%%%%%%%%% TITLE
\title{Probabilistic MIMO U-Net: Efficient and Accurate Uncertainty Estimation for Pixel-wise Regression}

\author{Anton Baumann, Thomas Roßberg, Michael Schmitt\\
University of the Bundeswehr Munich\\
Neubiberg, Germany\\
{\tt\small \{anton.baumann, thomas.rossberg, michael.schmitt\}@unibw.de}
}

\maketitle
\ificcvfinal\thispagestyle{empty}\fi

%%%%%%%%% ABSTRACT
\begin{abstract}
Uncertainty estimation in machine learning is paramount for enhancing the reliability and interpretability of predictive models, especially in high-stakes real-world scenarios. Despite the availability of numerous methods, they often pose a trade-off between the quality of uncertainty estimation and computational efficiency. Addressing this challenge, we present an adaptation of the Multiple-Input Multiple-Output (MIMO) framework -- an approach exploiting the overparameterization of deep neural networks -- for pixel-wise regression tasks. Our MIMO variant expands the applicability of the approach from simple image classification to broader computer vision domains. For that purpose, we adapted the U-Net architecture to train multiple subnetworks within a single  model, harnessing the overparameterization in deep neural networks. Additionally, we introduce a novel procedure for synchronizing subnetwork performance within the MIMO framework. Our comprehensive evaluations of the resulting MIMO U-Net on two orthogonal datasets demonstrate comparable accuracy to existing models, superior calibration on in-distribution data, robust out-of-distribution detection capabilities, and considerable improvements in parameter size and inference time. Code available at \href{https://github.com/antonbaumann/MIMO-Unet}{github.com/antonbaumann/MIMO-Unet}. 
\end{abstract}

%%%%%%%%% BODY TEXT
\section{Introduction}
\label{sec:intro}
Uncertainty estimation plays a crucial role in machine learning applications, enhancing the reliability and interpretability of predictive models. When deploying machine learning models in real-world scenarios, it is essential to be aware of the trustworthiness of their predictions in different contexts. Not only does uncertainty estimation give an indication of the model's confidence, but it also helps in the evaluation of the possible risks associated with the model's output.

A variety of methods have been proposed to address the challenge of uncertainty estimation in deep learning models. Among them are Monte Carlo dropout \cite{gal_dropout_2015, kendall2017uncertainties}, stochastic gradient Langevin dynamics \cite{welling2011bayesian}, Deep Ensembles \cite{lakshminarayanan2017simple,fort2019deep}, and Evidential Regression \cite{amini2020deep}. Each of these methods presents a delicate trade-off between the quality of uncertainty estimation and the computational costs or inference time associated with it.

The Multiple-Input Multiple-Output (MIMO) framework \cite{havasi_training_2021}, a new methodology exploiting the overparametrization of deep neural networks \cite{molchanov_pruning_2017}, exhibits considerable potential. This approach trains multiple subnetworks within a single network, thus permitting the exploration of numerous disconnected modes in weight space without necessitating an increase in parameters or causing inference delay~\cite{fort2019deep,havasi_training_2021}. By countering the main challenges associated with traditional ensemble models, the MIMO framework signifies a notable development towards enhancing the efficiency and robustness of uncertainty estimation methods.

\textbf{Key contributions:} The application of the MIMO framework until now has been limited to image classification. A significant gap exists in the application of MIMO for pixel-wise regression tasks common in Computer Vision and Earth Observation. In this paper, we adapt the MIMO approach for these tasks, thereby extending its benefits to a broader range of applications.

Additionally, we introduce a novel procedure for synchronizing subnetwork performance within the MIMO framework. This procedure is designed to ensure optimal functionality of the overall network by preventing individual subnetworks from either underperforming or overpowering the ensemble.

Finally, we undertake a comprehensive evaluation of our MIMO variant's ability to estimate epistemic uncertainty on regression tasks in both computer vision and earth observation domains. The performance of our MIMO model is contrasted with that of state-of-the-art neural network uncertainty estimation techniques. Through these comparisons, we aim to validate the efficacy and utility of MIMO U-Net for uncertainty estimation in machine learning.

\section{Related Work}
\label{sec:related_work}
Two main types of uncertainties often considered in predictions are \textit{aleatoric uncertainty} and \textit{epistemic uncertainty}. The former, arising from data randomness, is irreducible, while the latter, resulting from knowledge gaps, can diminish with more data or enhanced models~\cite{gawlikowski2021survey}. Aleatoric uncertainty is often addressed with Maximum a Posteriori (MAP) estimation~\cite{nix1994estimating}, while epistemic uncertainty is managed by employing a probability distribution over model weights.

However, obtaining the full posterior distribution over the parameters is computationally intractable~\cite{gawlikowski2021survey}. To approximate this distribution, several techniques have been introduced:

\textbf{Monte Carlo Dropout} (MC Dropout) \cite{gal_dropout_2015, kendall2017uncertainties} is a technique that approximates the posterior distribution over model weights in Bayesian Neural Networks. It repurposes dropout regularization \cite{srivastava2014dropout}, commonly employed to prevent overfitting during model training, as a tool for uncertainty estimation.

In the conventional use case, dropout is only implemented during the training phase. However, MC Dropout deviates from this approach by applying dropout in the testing or inference phase as well. This results in a multitude of "sub-models", each with a different dropout configuration, drawn from the approximate posterior.

It's noteworthy that MC Dropout inherently provides an approximation focused around a single mode of the distribution, capturing the peak uncertainty around the most probable model weights~\cite{fort2019deep}. Additionally, despite the need for multiple forward passes during inference, leading to slower prediction times, it is computationally efficient during training as only a single model needs to be trained.

\textbf{Deep Ensembles} \cite{lakshminarayanan2017simple} is another strategy for epistemic uncertainty estimation, leveraging multiple neural networks with varying weight initializations. This method effectively creates a diversified ensemble of independent deep models. Different initial weights, along with the randomness of stochastic gradient descent, facilitate slightly different input-output mappings for each model.

Each ensemble model offers unique predictions for a given input, collectively approximating the predictive posterior distribution. Deep Ensembles, unlike single-model approaches like MC Dropout, can explore multiple distribution modes \cite{fort2019deep}, providing a holistic uncertainty representation. However, this approach can be computationally costly both in terms of the increased number of parameters, due to multiple models, and inference delay caused by the need to generate and aggregate predictions from all models in the ensemble.

While ensembles have proven significantly effective, improvements to their performance are directed towards reducing computational time and the number of necessary parameters. Innovations such as Batch Ensemble \cite{wen_batchensemble_2020} and Bayesian Neural Networks with Rank-1 Factors \cite{dusenberry_efficient_2020} have optimized ensemble methods by sharing parameters across members, significantly cutting memory demands. Nonetheless, the continued need for multiple forward passes leaves room for further optimization.

\textbf{Evidential Regression}, as proposed by Amini et al. \cite{amini2020deep}, employs non-Bayesian neural networks to deliver estimations for both a target and its associated evidence, enabling the quantification of aleatoric and epistemic uncertainties. This is accomplished by setting evidential priors over the Gaussian likelihood function and instructing the network to infer the hyperparameters of the evidential distribution. As such, it offers a robust and efficient approach to uncertainty representation without the requirement for sampling at inference or training on out-of-distribution examples.
One of the key advantages of Evidential Regression is its computational efficiency, as it necessitates training only a single model and performing a single forward pass at inference. 
However, it is crucial to note that this method offers only a heuristic approximation of epistemic uncertainty, as pointed out by Meinert et al. \cite{meinert_unreasonable_2022}. Moreover, the method requires careful tuning of the regularization coefficient to ensure well-calibrated estimations \cite{amini2020deep}.
\newline\newline
A key insight drawn from past research indicates that a significant proportion of network parameters, specifically 70-80\%, can be pruned without significant impact on prediction performance \cite{molchanov_pruning_2017}. This finding suggests the possibility of latent capacity within networks that can be harnessed to train multiple subnetworks within a single network, a concept embodied in the Multiple-Input Multiple-Output (MIMO) method, recently introduced by Havasi et al.~\cite{havasi_training_2021}.

\textbf{The MIMO framework}~\cite{havasi_training_2021} leverages the over-parameterization inherent to deep neural networks \cite{frankle_lottery_2019, molchanov_pruning_2017}. These studies suggest that many neural network connections can be pruned without significant performance loss, indicating the possibility of multiple independent subnetworks within a single network. MIMO extends this concept by allowing these subnetworks, known as "winning tickets", to be trained concurrently without explicit separation \cite{havasi_training_2021}. This approach allows all subnetworks to be evaluated concurrently in a single forward pass during testing. This not only delivers the multi-mode exploration benefits typically offered by ensemble methods~\cite{havasi_training_2021}, but also provides the efficiency advantage of a single forward pass.

In the optimization process, MIMO employs stochastic gradient descent. During each training step, a subset $\{(\mathbf{x}_i, y_i)\}_{i \in B}$ of $|B|$ samples is drawn and randomly permuted by a permutation $\pi^{(i)}$ for each of the $m$ sub-models, ensuring the independence of model inputs. It is important to note that these permutations are different for each training step.
\begin{align}
    \left\{(\mathbf{x}_{\pi^{(1)}(i)}, y_{\pi^{(1)}(i)}), \dots, (\mathbf{x}_{\pi^{(m)}(i)}, y_{\pi^{(m)}(i)})\right\}_{i \in B}
\end{align}
At the input layer, the $m$ inputs $\bold{x}_{1:m} \coloneqq \bold{x}_{\pi^{(1)}(i)}, \dots, \bold{x}_{\pi^{(m)}(i)}$ are concatenated, and the network returns $m$ predictive distributions $p(y_1|\bold{x}_{1:m}, \boldsymbol{\theta}), \dots, p(y_m | \bold{x}_{1:m}, \boldsymbol{\theta})$ accordingly. 

The network is trained similarly to traditional neural networks, with the loss function being the average of the negative log-likelihoods of the predictions and a regularization term~\cite{loshchilov_decoupled_2019}.
\begin{align}
    \mathcal{L}(\boldsymbol{\theta}) = \frac{1}{m}\left[\sum^m_{i=1} -\log p(y_i|\bold{x}_{1:m}, \boldsymbol{\theta})\right] + R_\lambda(\boldsymbol{\theta})
\end{align}
During evaluation, the unseen input $\bold{x^*}$ is repeated $m$ times, s.t. $\bold{x^*}_{1:m} = \bold{x^*}, \dots, \bold{x^*}$, which independently approximates the predictive distribution 
\begin{align}
  p(y^*_i | \bold{x^*}, \dots, \bold{x^*}, \boldsymbol{\theta})\approx p(y^*, \bold{x^*}, \boldsymbol{\theta}), \;\; i = 1, \dots, m. \nonumber
\end{align}
The approximation of the predictive posterior distribution aligns with the methodologies utilized in MC Dropout~\cite{gal_dropout_2015} and Ensembles~\cite{lakshminarayanan2017simple}:
\begin{align}
p(y^*|\bold{x^*},\bold{x}_{1:n},y_{1:n}) \approx \frac{1}{m}\sum_{i=1}^m p(y^*_i, \bold{x^*}, \boldsymbol{\theta})
\end{align}
\textbf{Input repetition} The MIMO framework trains subnetworks using independent examples to prevent feature sharing. This works well with networks of ample capacity, but less so when capacity is limited. To address this, Havasi et al. \cite{havasi_training_2021} propose relaxing the independence between inputs. Rather than independently sampling $\bold{x}_1, \dots, \bold{x}_m$ from the training set, they may share the same value with a certain probability $\rho$. Specifically, $\bold{x}_1$ is sampled from the training set, and $\bold{x}_2, \dots, \bold{x}_m$ is set to be equal to $\bold{x}_1$ with probability $\rho$ or independently sampled with probability $1 - \rho$. This introduces a correlation in their joint distribution without affecting their marginal distributions.

\section{Adapting the MIMO Framework for Pixel-wise Regression}
In this study, we combine the Multiple-Input-Multiple-Output (MIMO) paradigm~\cite{havasi_training_2021} with the U-Net~\cite{ronneberger_u-net_2015} architecture, resulting in a novel approach for uncertainty-aware pixel-wise regression tasks. This fusion capitalizes on the innate capability of the U-Net architecture for image translation tasks, enriched with the inherent strengths of the MIMO framework in exploiting the over-parameterization of deep neural networks.

\subsection{MIMO U-Net}
\begin{figure}[h!]
\centering
\includegraphics[width=\linewidth]{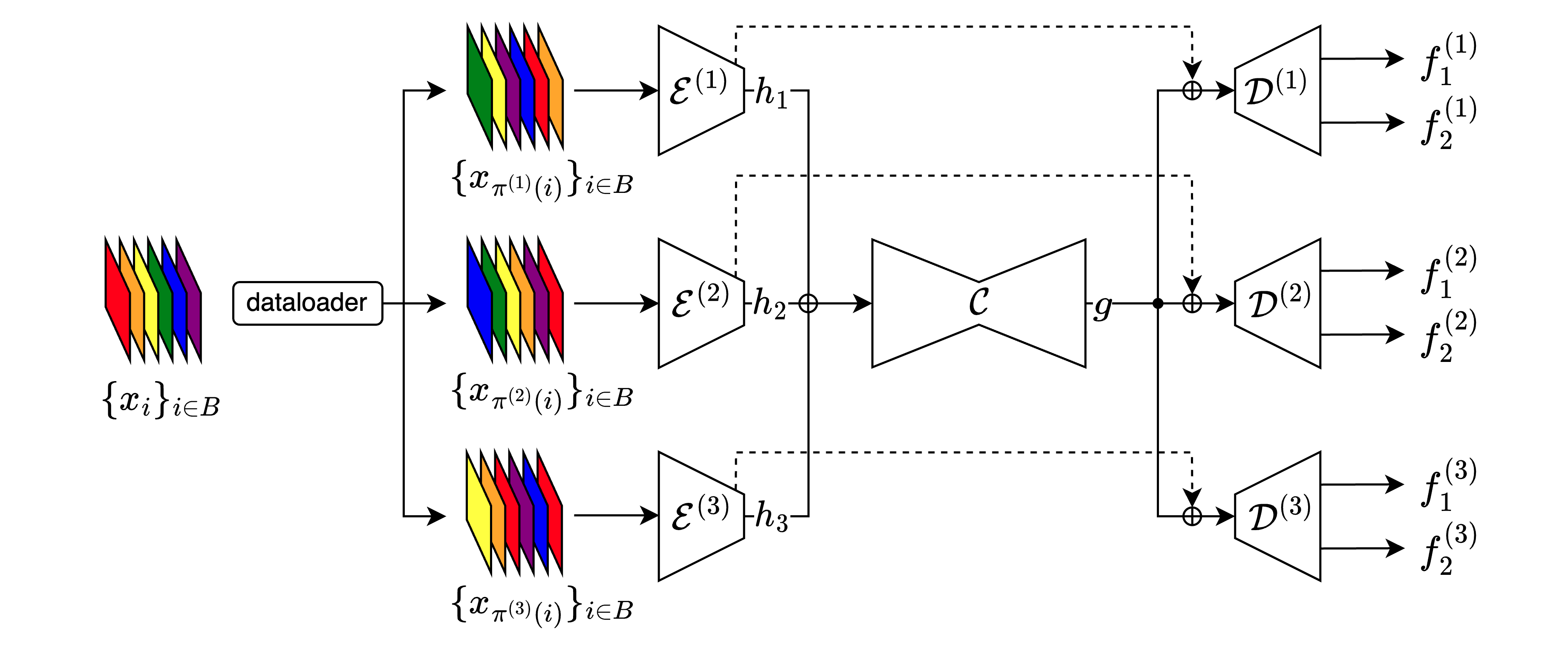}
\caption{MIMO U-Net architecture with three submodels. Dashed lines indicate skip-connections}
\label{fig:mimo_architecture}
\end{figure}

Our MIMO U-Net model (\autoref{fig:mimo_architecture}) comprises multiple distinct subnetworks, with each embodying an individual encoder-decoder pair $(\mathcal{E}^{(i)}, \mathcal{D}^{(i)})$ that is bridged by skip connections, a characteristic attribute of the U-Net architecture. The reasoning behind this design choice hinges on managing the potential overshadowing influence of skip connections. While skip connections are fundamental to the U-Net architecture, enabling low-level details to bypass the bottleneck and re-emerge in the decoder, they can inadvertently overpower the outcomes of other subnetworks. By implementing individual encoder-decoder pairs, we are able to prevent the overpowering influence of the skip connections from dominating the outcomes of the other subnetworks. This design decision helps us to maintain the autonomy of each subnetwork, which in turn promotes the creation of varied and distinct mappings from the input to the output images.

An essential note on the modifications introduced in our architecture relates to the initial and final layer of the traditional U-Net. In our MIMO U-Net model, the original first layer of the U-Net is replaced by the encoders of the subnetworks. Simultaneously, a similar restructuring occurs at the other end of the architecture, where the original U-Net's final layer is replaced by the decoders of the subnetworks. Despite this restructuring, we take care to ensure that the combined encoders and decoders of all subnetworks retain the same number of parameters as the first and last layers of the original U-Net. By doing so, we uphold the structural efficiency and computational economy of the U-Net architecture, all while instilling our model with the diversified, ensemble-like functionality that characterizes the MIMO framework.

In the first phase, each subnetwork processes the input image independently through its respective encoder, generating unique feature representations. 
\begin{align}
    \mathcal{E}^{(i)}(\bold{x}_i, \boldsymbol{\theta}) = h_i \:\:\:\text{for}\:\:\: i = 1, \dots, m
\end{align}
These representations are subsequently stacked and introduced to the core component of the U-Net architecture. This U-Net core $\mathcal{C}$, which maintains its own set of weights and parameters, generates a unified feature representation that is a confluence of the diverse learned features from all subnetworks.
\begin{align}
    \mathcal{C}(\text{stack}(h_1, \dots, h_m), \boldsymbol{\theta}) = g
\end{align}
The second phase involves the relay of this unified feature representation to each of the subnetworks' decoders. Each decoder, in turn, upscales the unified feature representations from the U-Net core and reconstructs the output image.
\begin{align}
    \mathcal{D}^{(i)}(g, \boldsymbol{\theta}) = \begin{bmatrix}
    f_1(\bold{x}_i, \boldsymbol{\theta})\\
    f_2(\bold{x}_i, \boldsymbol{\theta})
    \end{bmatrix} \:\:\:\text{for}\:\:\: i = 1, \dots, m
\end{align}
This design ensures that each subnetwork decoder receives the unified feature representation from the core. The distinct processing pathways that each subnetwork's encoder-decoder pair creates results in an array of diverse predictions. This ensemble-like behavior of the MIMO U-Net framework encapsulates model uncertainty and provides a broader spectrum of plausible pixel-wise regression maps for a given input.

\subsection{Training Criterion for Regression}
In regression problems, neural networks typically generate a single output, referred to as $\mu(\bold{x},\boldsymbol{\theta})$ and the parameters are optimized by minimizing the mean squared error (MSE) on the training set. However, this modelling choice does not allow us to capture predictive uncertainty. Following \cite{nix1994estimating}, we assume $\mathbf{y}_{i,j} | \bold{x}_i, \boldsymbol{\theta}$ to be Laplace distributed ($i$ specifying the subnetwork and $j$ the pixel index) and utilize the two outputs per submodule $f_1$ and $f_2$ to predict the distribution parameters:
\begin{align}
    \boldsymbol{\hat{\mu}}_i &\coloneqq \mu(\bold{x}_i,\boldsymbol{\theta}) \coloneqq f_1(\bold{x}_i,\boldsymbol{\theta})\\
    \boldsymbol{\hat{b}}_i &\coloneqq b(\bold{x}_i,\boldsymbol{\theta}) \coloneqq \exp(f_2(\bold{x}_i,\boldsymbol{\theta}))
    \label{eq:def_dist_params}
\end{align}
We opted for the Laplace distribution over the Gaussian likelihood because it typically outperforms L2 loss in vision \cite{kendall2017uncertainties} and provides improved differentiation of uncertainty for out-of-distribution data, as documented in \cite{nair_maximum_2022}.
With this, our model can be trained by optimizing
\begin{align}
    \mathcal{L}(\boldsymbol{\theta}) = \frac{1}{m}\sum^m_{i=1}\left[\frac{1}{d}\sum_{j=1}^d\log \boldsymbol{\hat{b}}_{i,j} + \frac{|\boldsymbol{y}_{i,j} - \boldsymbol{\hat\mu}_{i,j}|}{\boldsymbol{\hat{b}}_{i,j}}\right] + R_\lambda(\boldsymbol{\theta})
\end{align}
with the number of pixels $d$.
\subsection{Aleatoric and Epistemic Variance}
Using the predictions $\boldsymbol{\hat{\mu}}_i$ and $\boldsymbol{\hat{b}}_i$ as described in \eqref{eq:def_dist_params}, we can estimate the mean of our predictive posterior as follows:
\begin{align}
    \boldsymbol{\Bar{\mu}}(\bold{x^*}) \coloneqq \EX[\boldsymbol{y}^*|\bold{x^*}, \bold{x}_{1:n}, \bold{y}_{1:n}] \approx \frac{1}{m}\sum^m_{i=1}\boldsymbol{\hat{\mu}}_i
    \label{eq:mean_estimation}
\end{align}
Furthermore, by equating $\boldsymbol{\hat{\sigma}}^2_{i,j} = 2\boldsymbol{\hat{b}}^2_{i,j}$~\cite{kotz_laplace_2001}, we can estimate the \textit{aleatoric variance} and \textit{epistemic variance} in our predictions \cite{kendall2017uncertainties}:
\begin{align}
    &\Var\left[\bold{y}^*|\bold{x^*},\bold{x}_{1:n},\bold{y}_{1:n}\right] \nonumber\\
    &=  \underbrace{\frac{1}{m}\sum^m_{i=1}\boldsymbol{\hat{\sigma}}^2_{i}}_{\text{aleatoric uncertainty}} + \underbrace{\frac{1}{m-1}\sum^m_{i=1}\left[\boldsymbol{\hat{\mu}}_i - \boldsymbol{\Bar{\mu}}(\bold{x^*})\right]}_{\text{epistemic uncertainty}}
    \label{eq:variance_decompositions}
\end{align}

\subsection{Submodel Synchronization}
\begin{figure}[h!]
\centering
\includegraphics[width=0.9\linewidth]{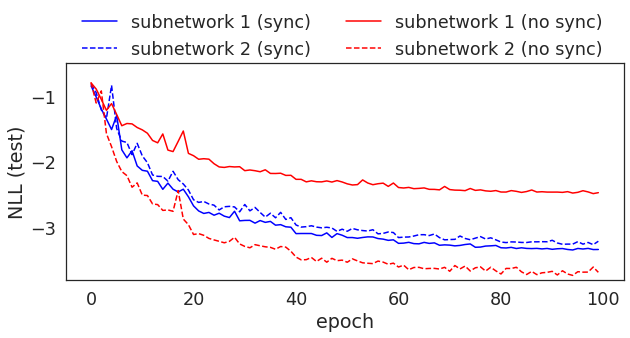}
\caption{The negative log-likelihood of the test set for each epoch. The learning of individual subnetworks is uneven without subnetwork synchronization.}
\label{fig:loss_synchronization_line_plot}
\end{figure}

A key aspect of training multiple subnetworks within a single network, as in the MIMO framework, is the inherent diversity in learning trajectories across subnetworks. This diversity, influenced by differences in initialization and gradient updates, often leads to subnetworks learning at varied rates. The manifestation of this phenomenon is clear in Figure \ref{fig:loss_synchronization_line_plot}, where loss trajectories with and without subnetwork synchronization are depicted. In the absence of synchronization, the progression of learning rates is distinctly uneven.
Recognizing this variability as an opportunity for improvement, we introduce a synchronization mechanism that harmonizes the learning across subnetworks. This mechanism monitors the losses of the last $k$ training steps for each subnetwork and assigns weights to the current loss, thereby facilitating synchronization in training progress. The weight for each of the $m$ submodels, denoted as $\bold{w}_i$, is determined by:

\begin{equation}
\bold{w}_i = \frac{m\exp{(\nicefrac{\Bar{l}_i}{\tau}})}{\sum^m_{j=1}\exp{(\nicefrac{\Bar{l}_j}{\tau}})} 
\end{equation}

In this formulation, $\Bar{l}_i$ denotes the mean losses of the last $k$ steps for submodel $i$, while $\tau$ regulates the weight concentration through a temperature parameter. Multiplying the softmax result by $m$ maintains the overall learning rate.

\section{Experiments}
We present our experiments applying pixelwise regression to two orthogonal tasks: \textit{monocular depth estimation}, representing high-resolution, horizontally captured photos in computer vision, and \textit{NDVI score prediction} in earth observation, using overhead Synthetic Aperture Radar (SAR) data with radar backscatter measurements. The datasets used for these tasks are NYU Depth v2 \cite{silberman_indoor_2012} and SEN12TP \cite{rosberg_globally_2023}, respectively.
The orthogonality of these tasks lies in their distinct nature and operational environments. Monocular depth estimation involves processing traditional, high-resolution photography, while NDVI score prediction deals with SAR data, exhibiting vastly different data characteristics and challenges.

We test the scalability and versatility of MIMO U-Net against commonly used methods like Deep Ensembles \cite{lakshminarayanan2017simple}, MC Dropout \cite{gal_dropout_2015}, and Deep Evidential Regression~\cite{amini2020deep} under various sample frequencies $m$. Performance on unseen data, both in-distribution and out-of-distribution, is evaluated based on accuracy and predictive uncertainty. Metrics reported include MAE, RMSE, NLL, ECE \cite{kuleshov_accurate_2018}, number of parameters, and inference speed.

\subsection{Monocular Depth Estimation}
\label{subsec:monocular_depth_estimation}

\begin{figure}[h]
\centering
\includegraphics[height=110px]{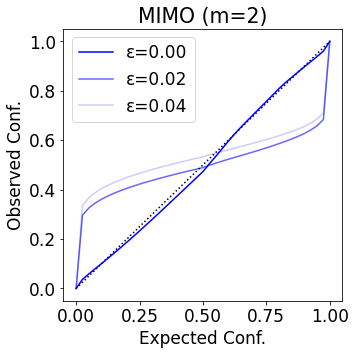}\hfill
\includegraphics[height=110px]{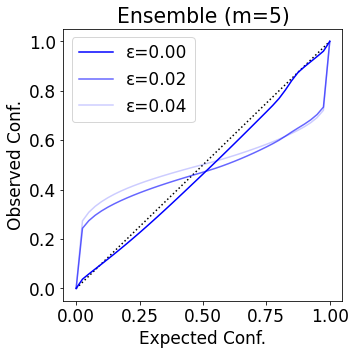}
\includegraphics[height=110px]{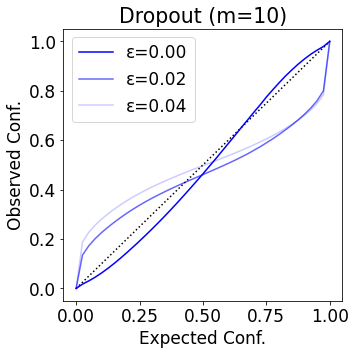}\hfill
\includegraphics[height=110px]{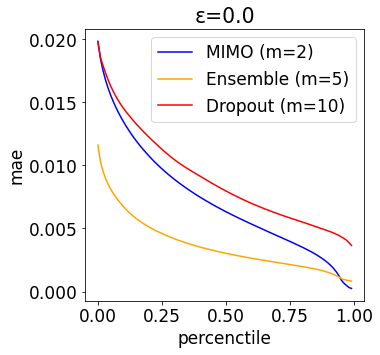}
\caption{Calibration plots \cite{kuleshov_accurate_2018} for various adversarial perturbations intensities $\epsilon$ using the Fast Gradient Sign Method. MIMO has best calibration on unperturbed data.
\textbf{Bottom right:} Precision-recall plot for different methods on unperturbed data}
\label{fig:precision_recall_and_calibration}
\end{figure}

In our initial experiments, we used MIMO U-Net for pixelwise regression on the NYU Depth v2 dataset for depth estimation. This dataset by Silberman et al.~\cite{silberman_indoor_2012} includes 27k+ RGB image samples from various indoor scenes with associated depth maps (160, 128). Dataset details and preprocessing can be found in \cite{amini2020deep}. The ApolloScape dataset~\cite{huang_apolloscape_2018} was used to assess our model's out-of-distribution detection.

In training, the MIMO U-Net had 42 base channels, split across two subnetworks. A single subnetwork with 0.1 dropout probability mirrored the U-Net architecture for spatial dropout. For the ensemble setup, five unique U-Net models were used, all trained without dropout. For Evidential Regression, we utilized the PyTorch~\cite{Paszke_PyTorch_An_Imperative_2019} implementation of the loss function from \cite{soleimany2021evidential}.
All models were trained over 100 epochs, utilizing the Adam optimizer with a learning rate of $1e^{-4}$ and a decay policy in place. The learning rate decay was characterized by a gamma value of 0.5 and a step size of 20. For computing predictions and uncertainty estimations, we used \autoref{eq:mean_estimation} and \autoref{eq:variance_decompositions} respectively. The subnetworks were synchronized by tracking the $k=10$ most recent loss values and setting the temperature $\tau = 0.3$.

A comparative review of the metrics reveals that the MIMO U-Net model, without input repetition, achieves results comparable to the MC Dropout model in terms of both MAE and RMSE. Furthermore, the MIMO U-Net model exhibits a better NLL than the dropout model and nearly perfect calibration, as indicated by the ECE score and the calibration plot in \autoref{fig:precision_recall_and_calibration}.

A crucial advantage of the MIMO U-Net over the MC Dropout and Ensemble method is the inference speed. Due to the structure of MIMO U-Net, it requires only a single forward pass to generate predictions, significantly enhancing computational efficiency while maintaining an equal number of parameters to the MC Dropout model.

The results of this experimental setup are summarized in \autoref{tab:nyuv2_depth_results} and \autoref{fig:precision_recall_and_calibration}. 
\begin{table}[h!]
  \centering
  \resizebox{\linewidth}{!}{%
  \begin{tabular}{@{}lrrrrrrr@{}}
    \toprule
    \textbf{model} & m & \#Params & Inf. Speed & $\downarrow$ MAE & $\downarrow$ RMSE & $\downarrow$ NLL & $\downarrow$ ECE \\
    \midrule
    % https://wandb.ai/anton-baumann/MIMO%20NYUv2Depth%20M/runs/py4yps4w
    MIMO U-Net & 2 & \underline{7.38 M} & \underline{4 ms} & 0.020 & 0.041 & -3.386 & \underline{0.008} \\
    \midrule
    MC Dropout & 5 & 7.44 M & 14 ms  & 0.020 & 0.038 & -3.133 & 0.047 \\
    MC Dropout & 10 & 7.44 M & 27 ms & 0.019 & 0.037 & -3.134 & 0.078 \\
    \midrule
    Ensembles & 5 & 37.20 M & 14 ms & \underline{0.012} & \underline{0.031} & \underline{-4.040} & 0.030 \\
    \midrule
    Evidential & 1 & 7.44 M & \underline{4 ms} & 0.019 & 0.037 & -3.165 & 0.047 \\
    \bottomrule
  \end{tabular}}
  \caption{NYU Depth v2 dataset results: $m$ denotes submodule count for MIMO or sample count for Dropout, Ensembles.}
  \label{tab:nyuv2_depth_results}
\end{table}

\paragraph{Submodel Synchronization}
The effectiveness of submodel synchronization was evaluated by performing an ablation study on the synchronization parameters, specifically the temperature ($\tau$) and the number of steps ($k$). Optimal performance was identified at $\tau = 0.3$ and $k = 10$. Table \ref{tab:submodule_sync_effect} clearly illustrates the substantial improvements in all metrics when the synchronization process is utilized.

\begin{table}[h!]
\centering
\begin{tabular}{@{}lcrrrr@{}}
\hline\noalign{\smallskip}
dataset & sync. & $\downarrow$ MAE & $\downarrow$ RMSE & $\downarrow$ NLL & $\downarrow$ ECE \\
\hline\noalign{\smallskip}
NYUv2 & \textbf{no} & 0.026 & 0.047 & -3.021 & 0.029 \\
NYUv2 & \textbf{yes} & \underline{0.020} & \underline{0.041} & \underline{-3.386} & \underline{0.008} \\
\hline\noalign{\smallskip}
\end{tabular}
\caption{Results on NYU Depth v2 dataset with and without submodel synchronization.}
\label{tab:submodule_sync_effect}
\end{table}

\paragraph{Assessment of Out-of-Distribution Testing}
A robust machine learning model must discern out-of-distribution (OOD) test data, deviating from training instances—a key facet of model uncertainty. This study evaluates the MIMO U-Net model's capacity to amplify predictive uncertainty upon encountering OOD data.

Our experimentation was carried out using two distinct datasets. The first, the NYU Depth v2 test set \cite{silberman_indoor_2012}, serves as an in-distribution (ID) data representative, consisting of instances the model should be familiar with, as it was trained on data from the same distribution. The second dataset, the ApolloScape dataset~\cite{huang_apolloscape_2018}, containing a diverse range of complex and realistic driving scenarios, was chosen for its contrasting properties to the ID data and represents the OOD data. 

\begin{figure}[h!]
\centering
\includegraphics[width=\linewidth]{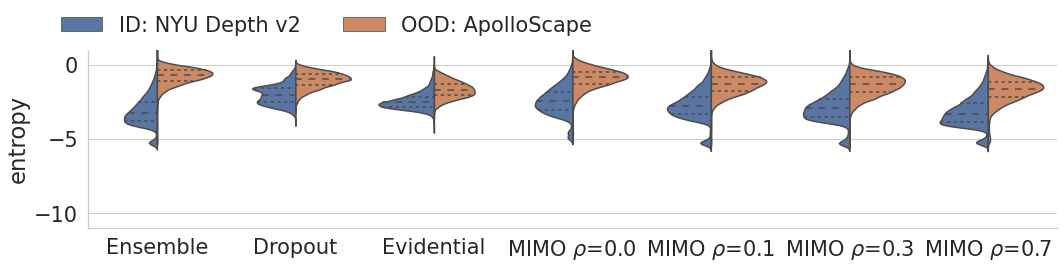}
\caption{Distribution of \textit{combined entropy}: All methods estimate low uncertainty (entropy) on in-distribution (ID) data and inflate uncertainty on out-of-distribution (OOD) data.}
\label{fig:method_comparison_id_ood}
\end{figure}

\begin{figure}[h]
\centering
\includegraphics[width=\linewidth]{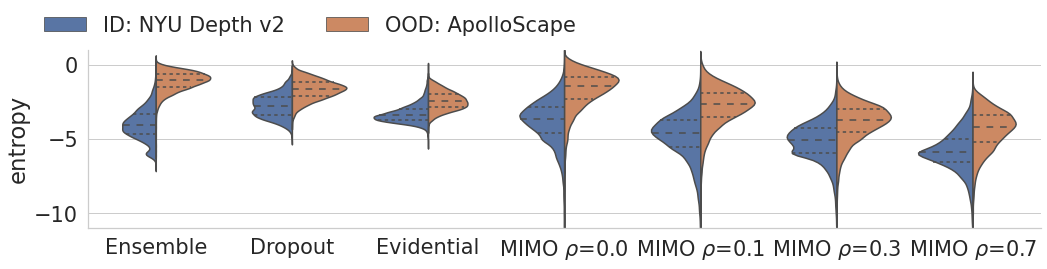}
\caption{\textit{Epistemic entropy} distribution: Input repetition probability increase leads to a decrease in subnetworks prediction diversity.}
\label{fig:effect_input_rep_on_epistemic_unc}
\end{figure}

Given the contrast between the two datasets, we expected the model to demonstrate a higher degree of uncertainty when encountering the OOD data. Our hypothesis was corroborated by the following empirical results. A notable increase (comparable to the behavior of the ensemble) in the level of uncertainty can be observed when the model is exposed to the OOD data. Particularly, as depicted in \autoref{fig:effect_input_rep_on_epistemic_unc}, Evidential Regression exhibits only modest differentiation between ID and OOD. This observation aligns with the discoveries detailed in \cite{nair_maximum_2022}. \autoref{fig:method_comparison_id_ood}, which represents entropy distributions of the total model uncertainty, illustrates this shift in uncertainty between ID and OOD data.

\begin{figure*}[h!]
\centering
\includegraphics[width=\textwidth]{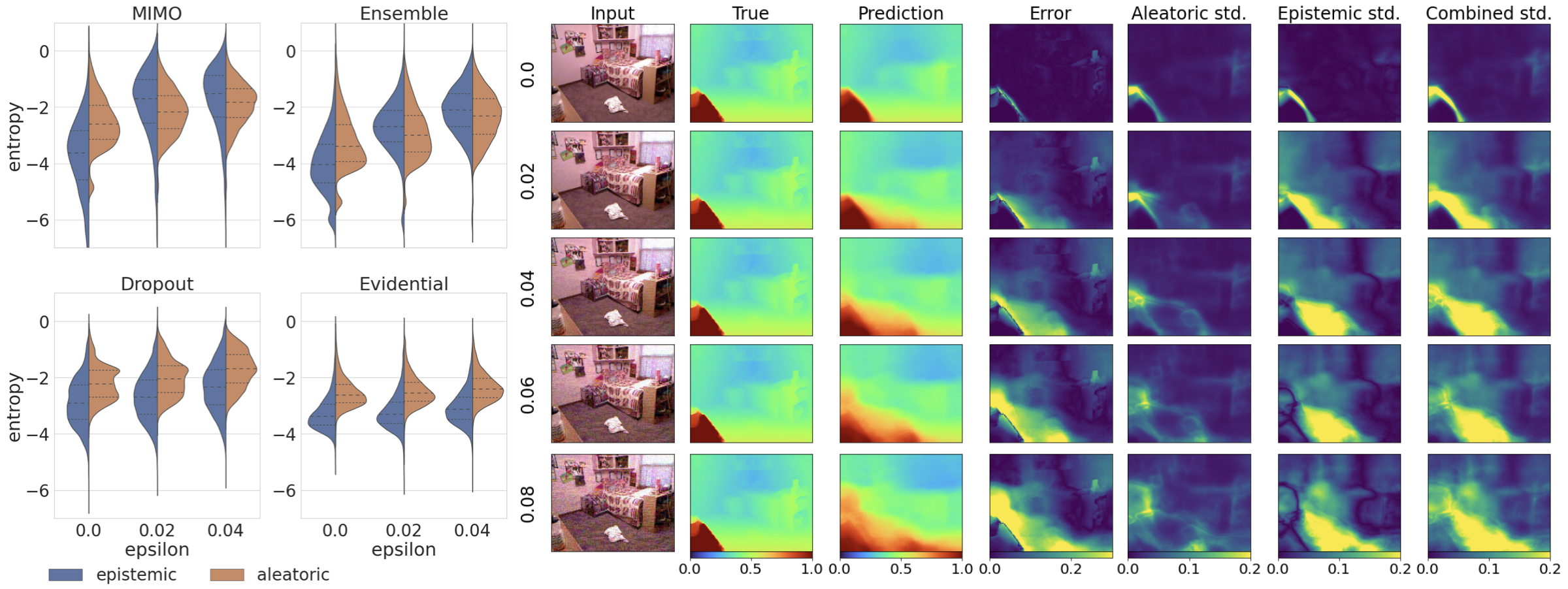}
\caption{\textbf{Left:} Entropy distributions for methods against rising adversarial perturbations $\epsilon$. \textbf{Right:} MIMO U-Net's test prediction with increasing $\epsilon$. MIMO shows resilience to adversarial noise, with its uncertainty estimates mirroring prediction error. Significant epistemic entropy shifts occur in MIMO and Ensemble under amplified perturbation.}
\label{fig:mimo_adversarial_grid}
\end{figure*}

\paragraph{Robustness to Adversarial Samples}
Here, we focussed on Out-Of-Distribution detection when inputs are adversarially perturbed using the Fast Gradient Sign Method (FGSM) \cite{goodfellow_explaining_2015}. 
As shown in \autoref{fig:mimo_adversarial_grid}, increasing adversarial perturbation parameter~$\epsilon$ shifts the MIMO U-Net's epistemic entropy distribution and amplifies MIMO's prediction error (0.01, 0.08, 0.11 for $\epsilon$ = 0, 0.02, 0.04), correlating with rising uncertainty. Here, the predictive epistemic uncertainty rises with the increase in noise, while maintaining a strong correlation with the error in terms of the spatial distribution within the image. Evidential Regression's increasing MAE (0.02, 0.07, 0.09) shows no marked shift in predictive uncertainties.

\paragraph{Effect of Input Repetition on Epistemic Uncertainty}
\label{subsubsec:effect_of_input_rep}
As we increase the input repetition probability $\rho$, the MIMO U-Net's performance in terms of MAE, RMSE, and NLL gradually approaches the performance of the ensemble (cf.~\autoref{tab:nyuv2_depth_ablation}). However, it's important to note that as $\rho$ increases, the variance in the model's predictions decreases, which results in the underestimation of the model's epistemic uncertainty (cf.~\autoref{fig:effect_input_rep_on_epistemic_unc}). This suggests that higher probabilities of input repetition are likely to degrade the calibration on OOD data.

\begin{figure}[h]
\centering
\includegraphics[width=\linewidth]{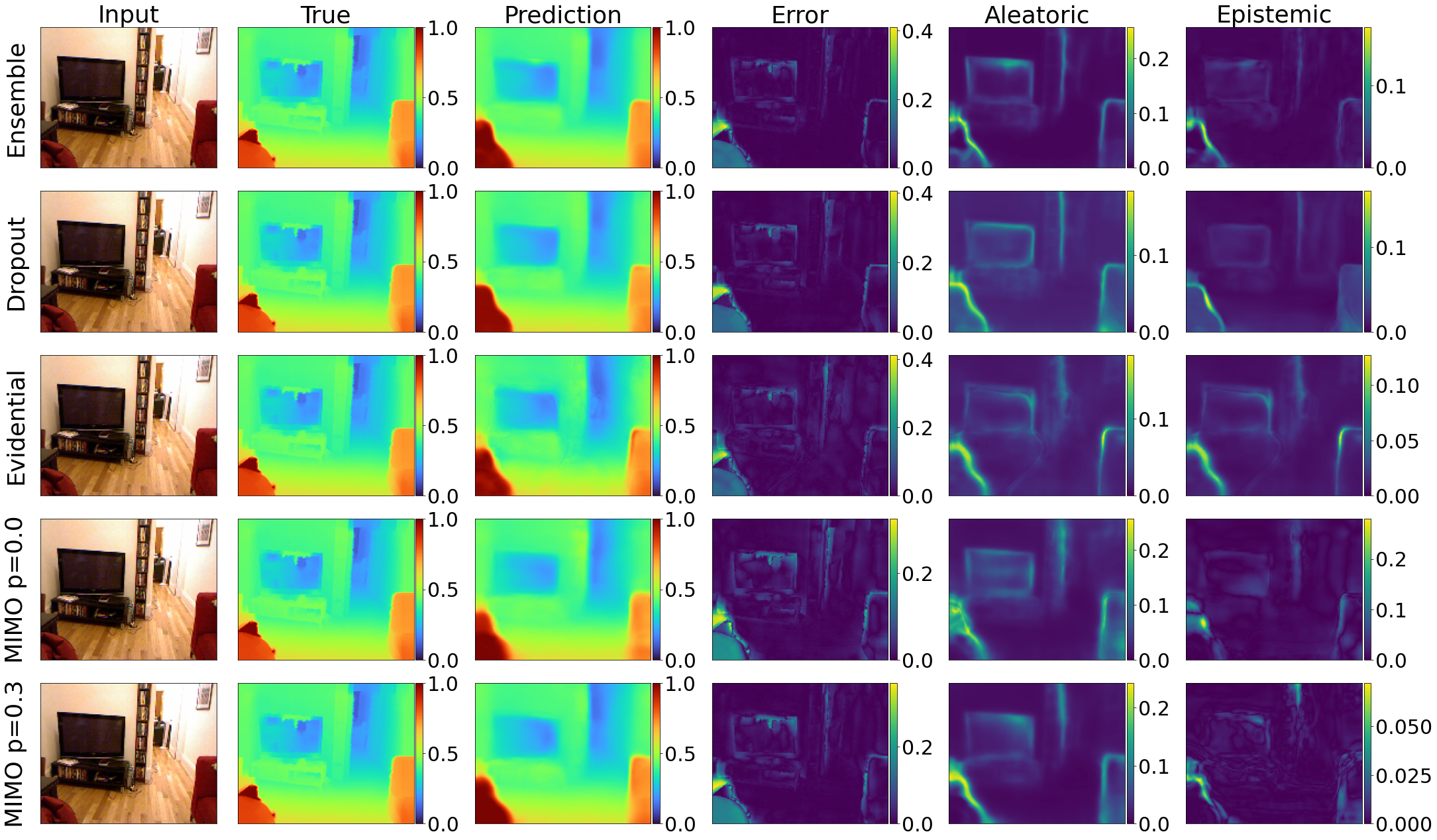}
\caption{Epistemic Uncertainty estimation semantically aligns with ensemble and dropout estimations}
\label{fig:image_plots_uncertainty}
\end{figure}

Despite this, as visualized in \autoref{fig:image_plots_uncertainty}, the MIMO approach still manifests significant epistemic uncertainties at identical positions to the ensemble method. Intriguingly, when subjected to high input repetitions, the resultant epistemic uncertainty remains semantically sensible, hinting at the possibility of calibrating epistemic uncertainties for future research.

\begin{table}[h!]
  \centering
  \resizebox{\linewidth}{!}{%
  \begin{tabular}{@{}lrrrrrr@{}}
    \toprule
    \textbf{model} & $\rho$ & m & $\downarrow$ MAE & $\downarrow$ RMSE & $\downarrow$ NLL & $\downarrow$ ECE \\
    \midrule
    % https://wandb.ai/anton-baumann/MIMO%20NYUv2Depth%20M/runs/py4yps4w
    MIMO U-Net & 0.0 & 2 & 0.020 & 0.041 & -3.386 & \underline{0.008} \\
    % https://wandb.ai/anton-baumann/MIMO%20NYUv2Depth%20M/runs/28kad7a1
    MIMO U-Net & 0.1 & 2 & 0.018 & 0.040 & -3.542 & 0.009 \\
    MIMO U-Net & 0.3 & 2 & 0.017 & 0.038 & -3.593 & 0.056 \\
    MIMO U-Net & 0.5 & 2 & 0.015 & 0.036 & -3.696 & 0.044  \\
    MIMO U-Net & 0.7 & 2 & \underline{0.014} & \underline{0.034} & \underline{-3.805} & 0.047 \\
    \bottomrule
  \end{tabular}}
  \caption{Results on NYU Depth v2 dataset: $\rho$ denotes the input repetition probability and $m$ the submodel count.}
  \label{tab:nyuv2_depth_ablation}
\end{table}

\subsection{NDVI Estimation from Radar Backscatter}
\begin{figure}[b]
    \centering
    \includegraphics[height=85px]{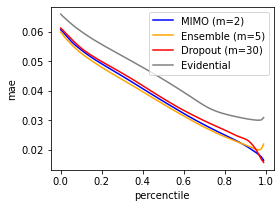}
    \includegraphics[height=85px]{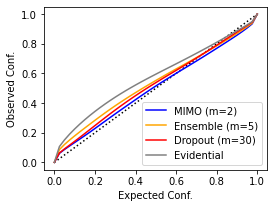}
    \caption{\textbf{Left:} Precision-recall plot for different methods on unseen test data. \textbf{Right:} Calibration plot: all method show almost perfect calibration
}
    \label{fig:precision_recall+cal_ndvi}
\end{figure}

In earth observation, the normalized difference vegetation index (NDVI) is commonly used for vegetation monitoring.
It is calculated from red ($R$) and infrared ($IR$) spectral bands of optical data using $NDVI=\frac{(IR-R)}{(IR+R)}$ and represents vegetation health as a scalar value in $[-1, 1]$.
As the NDVI relies on optical data, clouds pose a problem because they obstruct the view of the earth's surface from space.
One solution to the problem of cloud coverage is to use synthetic aperture radar (SAR) data, which can penetrate clouds.
However, the vastly different data characteristics require a complex, nonlinear transformation of SAR backscatter into NDVI values.

To translate SAR images to NDVI images, we use the SEN12TP dataset \cite{rosberg_globally_2023} and perform a pixelwise regression of the two channels of the SAR data (resulting from receiving the backscattered signal at two different polarizations) to the NDVI values derived from an optical sensor. 
Training our models, we followed earlier configurations (\ref{subsec:monocular_depth_estimation}), but limited the training period to 40 epochs and employed a MIMO U-Net model with 60 base channels.

In our assessment of MIMO U-Net, MC Dropout, Ensembles, and Evidential Regression for the outlined task, the MIMO U-Net model emerged as notably effective in its performance. The MAE, RMSE, and NLL of each method were similar, with MIMO U-Net outperforming in terms of inference speed and presenting the lowest Expected Calibration Error (ECE) for 2 and 4 subnetworks. Notably, in relation to inference speed and parameter count, these findings are consistent with those obtained from the NYU Depth v2 dataset (see \autoref{tab:nyuv2_depth_results}).

\begin{table}[h!]
\centering
\resizebox{\linewidth}{!}{%
\begin{tabular}{@{}lrrrrrrr@{}}
    \toprule
    \textbf{model} & m & $\downarrow$ \#Params & Inf. Speed & $\downarrow$ MAE & $\downarrow$ RMSE & $\downarrow$ NLL & $\downarrow$ ECE \\
    \midrule
    MIMO U-Net & 2 & 15.06 M & \underline{11 ms} & 0.121 & 0.179 & -1.236 & 0.030 \\
    MIMO U-Net & 3 & 15.04 M & 12 ms & 0.124 & 0.181 & -1.202 & 0.110 \\
    MIMO U-Net & 4 & \underline{15.03 M} & 15 ms & 0.129 & 0.186 & -1.147 & \underline{0.023} \\
    \midrule
    MC Dropout &  2 & 15.19 M & 20 ms & 0.124 & 0.180 & -1.199 & 0.075 \\
    MC Dropout & 10 & 15.19 M & 65 ms & 0.122 & 0.177 & -1.219 & 0.067  \\
    MC Dropout & 30 & 15.19 M & 171 ms & 0.122 & 0.178 & -1.216 & 0.051  \\
    \midrule
    Ensembles & 2 & 30.38 M & 20 ms & 0.122 & 0.179 & -1.220 & 0.153 \\
    Ensembles & 5 & 75.88 M & 41 ms & \underline{0.120} & \underline{0.175} & \underline{-1.248} & 0.175 \\
    \midrule
    Evidential & 1 & 15.19 M & \underline{11 ms} & 0.132 & 0.187 & -1.102 & 0.352 \\
    \bottomrule
  \end{tabular}}
\caption{Results on SEN12TP dataset}
\label{tab:ndvi_results}
\end{table}

\begin{figure}[htb]
    \centering
    % Notebook: thomas - ~/dev/2023_uncertainty_analysis/MIMO_NDVI.ipynb
    % data from mimo2_seed1, idx=1984 
    \includegraphics[width=\columnwidth]{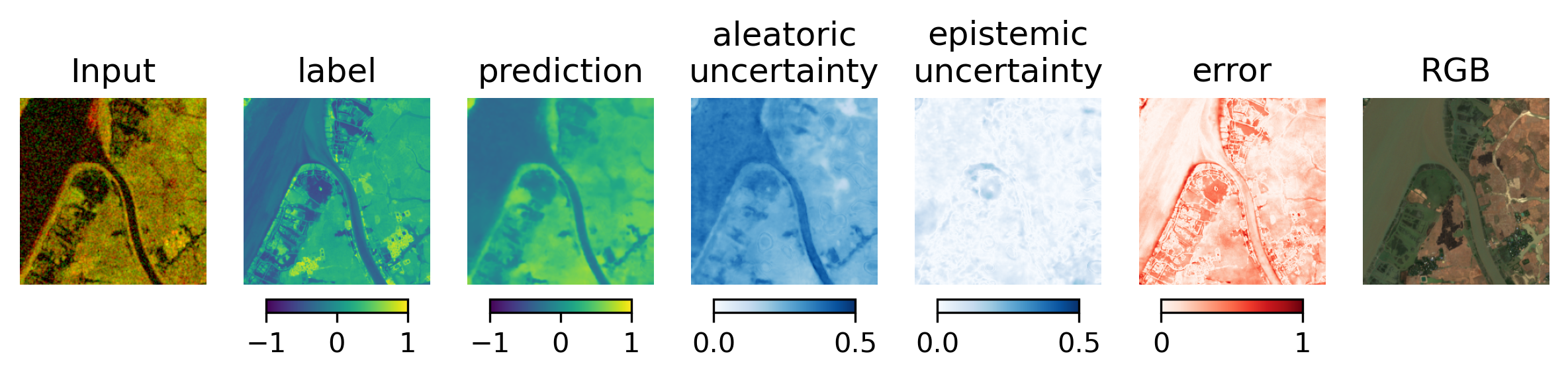}
    \caption{Results for one area with a river contained in the top-left corner. 
    For the water surface, a high aleatoric uncertainty can be seen, whereas the epistemic uncertainty is fairly homogeneous. 
    %Only pixels from the first three classes (water, wetland, ice) were included during masked training.
    }
    \label{fig:uncertainty-over-water}
\end{figure}

Our trained model highlights various influences on error and uncertainty. Of particular note, water surfaces generate an aleatoric variance of \qty{0.103}, which substantially surpasses the \qty{0.036} from non-water surfaces (\autoref{tab:comparision_uncertainty_water}). This discrepancy originates from the electromagnetic spectrum properties unique to SAR sensors: their emitted radar signals, incapable of water penetration, are reflected at the surface, thereby excluding under-water features from the captured data. Conversely, visible light, used by optical sensors, can penetrate water to some extent, allowing data acquisition from the topmost water layers. Therefore, the model's inability to learn an accurate mapping between SAR inputs and outputs results in heightened aleatoric uncertainty for aquatic regions, as exemplified by the river in~\autoref{fig:uncertainty-over-water}.

\begin{table}[]
    \centering
    \sisetup{round-mode=places,round-precision=3}
    \resizebox{\linewidth}{!}{%
    \begin{tabular}{lSSS[round-precision=4]S}
    \toprule
    {} &     {error} &  {aleatoric var} &  {epistemic var} &  {combined var} \\
    \midrule
    water       & 0.272006 & 0.102668 & 0.003447 & 0.106114 \\
    non water   & 0.117216 & 0.035561 & 0.002382 & 0.037943 \\
    \bottomrule
    \end{tabular}}
    \caption{Comparison of error and uncertainty for different water and non-water land cover surfaces.}
    \label{tab:comparision_uncertainty_water}
\end{table}

\begin{figure}[htb]
    \centering
    \includegraphics[width=\columnwidth]{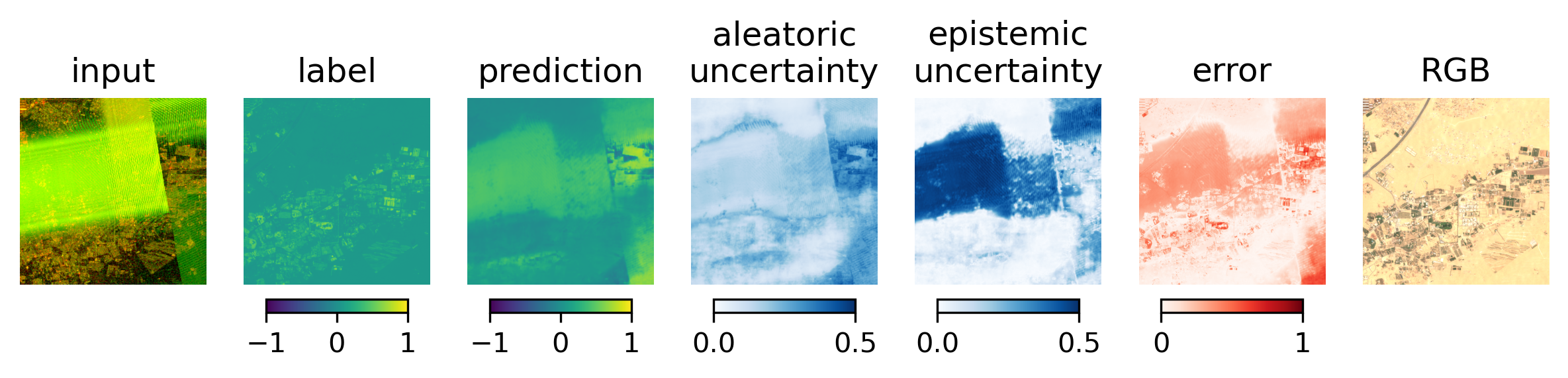}
    \caption{Radio wave interferences result in artifacts in the SAR image (bright yellow-greenish squares).
    This results in a high error and epistemic uncertainty of the model prediction.
    }
    \label{fig:corrupted-sar-data}
\end{figure}

SAR imagery can occasionally exhibit artifacts resulting from radio frequency interferences. These interferences occur when the SAR sensor picks up signals from other transmitters operating within the same frequency band. An exemplar case was observed in SAR data acquired over Dubai (United Arab Emirates) on 8th July 2019, which showed significant corruption due to these interferences. Upon applying our model to this affected data, we noticed an elevated epistemic uncertainty across all regions presenting artifacts. This pronounced increase in epistemic uncertainty across artifact-affected regions underscores the model's practical utility in detecting out-of-distribution data, as depicted in \autoref{fig:corrupted-sar-data}.

\begin{figure}
    \centering
    \includegraphics[width=\columnwidth]{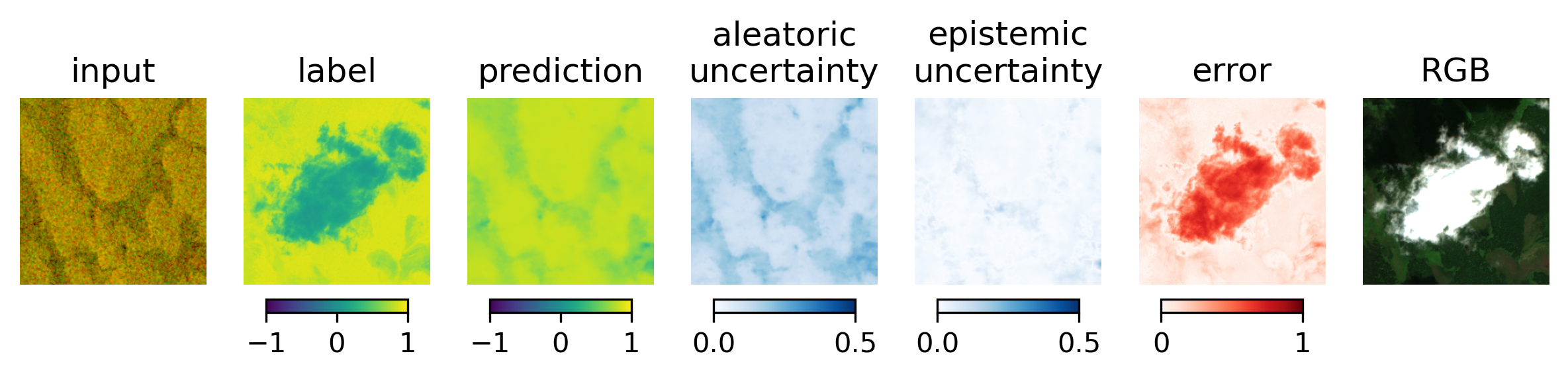}
    \caption{Image area where the optical data is cloud-contaminated, resulting in misleading NDVI values.
    This results in a high error between optical and predicted NDVI, even though the model predicts a low uncertainty.
    }
    \label{fig:uncertainty-for-cloud}
\end{figure}

Clouds in the optical images result in errors in our use case because the vegetation on the ground is occluded.
When the model prediction is compared to cloud-contaminated optical imagery, a high error between predicted and optical NDVI occurs.
The model still predicts a low uncertainty for these areas because clouds do not change what information is acquired by the SAR sensor.
This is apparent in \autoref{fig:uncertainty-for-cloud}.

\begin{figure}
    \centering
    \includegraphics[width=\columnwidth]{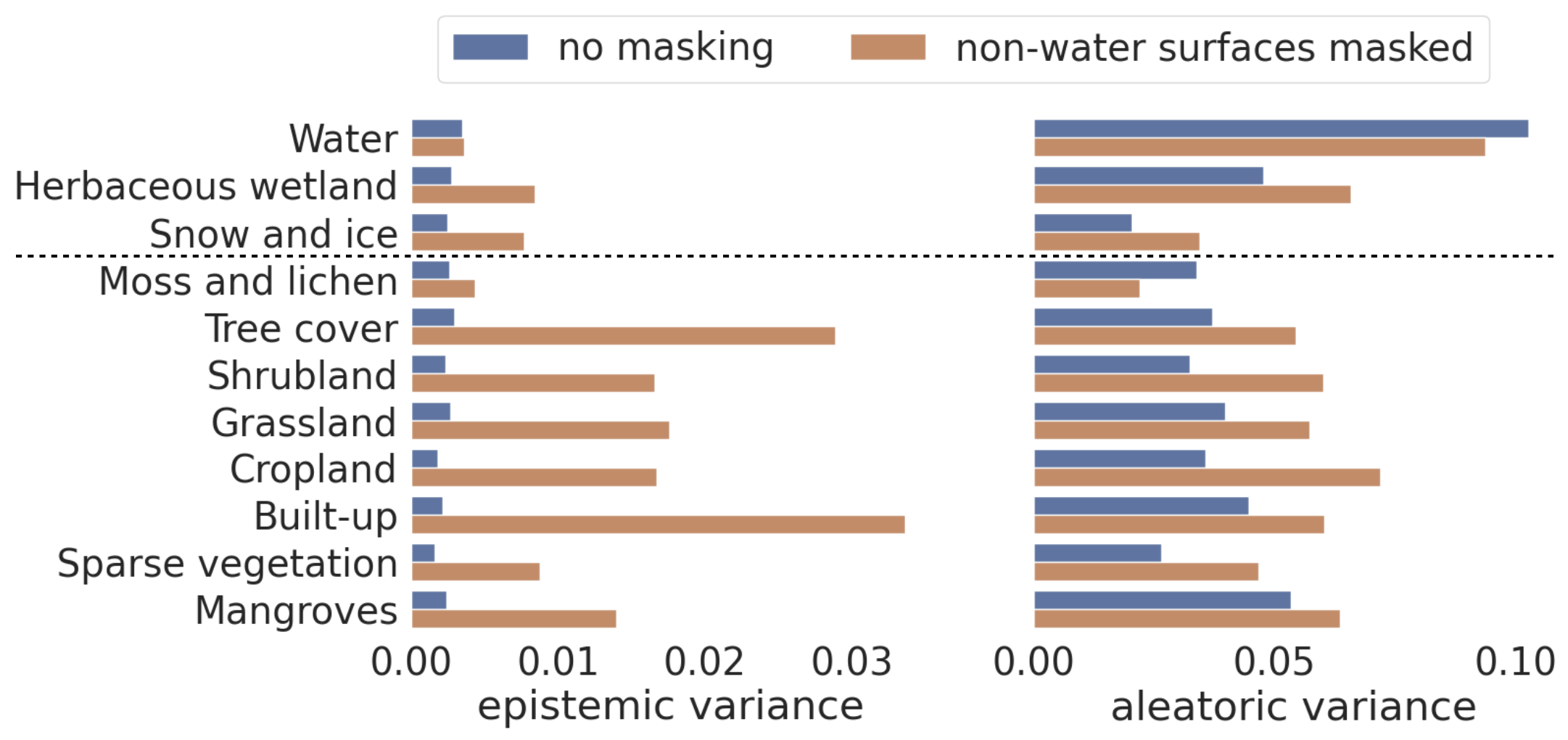}
    \caption{Significantly higher epistemic uncertainty on landcover classes that were excluded during training. Only pixels from the first three classes (water, wetland, ice) were included.}
    \label{fig:comparision-landcover-masked}
\end{figure}

Finally, to investigate the behavior of our model with respect to unseen landcover classes, we conducted an additional experiment where we masked all surfaces except for water, herbaceous wetland, and ice during training. Upon evaluation and comparison with the model trained on the complete SEN12TP dataset, we observed a pronounced increase in epistemic variance in the masked landcover classes for the landcover-masked model. The divergence in aleatoric variance between both models was less significant, as depicted in \autoref{fig:comparision-landcover-masked}.

\section{Summary \& Conclusion}
In this study, we successfully expanded the scope of the MIMO approach by adapting it for pixel-wise regression tasks, thereby broadening its relevance in the computer vision domain. A novel submodel synchronization procedure was introduced as a key contribution, optimized to maintain the balance within the overarching network by mitigating the possibility of any individual subnetwork underperforming or overpowering the ensemble. We validated this approach with exhaustive evaluations on the MIMO U-Net using two orthogonal datasets from computer vision and earth observation fields. Our empirical results revealed comparable accuracy to established models, exceptional calibration on in-distribution data, and convincing demonstration of its robust out-of-distribution detection capabilities, all while exhibiting significant improvements in parameter size and inference time. Therefore, this study paves the way for the implementation of MIMO in the broader computer vision realm, exhibiting potential for resource-optimized, high-accuracy tasks while maintaining a desirable trade-off between performance and computational load.

{\small
\bibliographystyle{ieee_fullname}
\bibliography{egbib}

\begin{thebibliography}{10}\itemsep=-1pt

\bibitem{amini2020deep}
Alexander Amini, Wilko Schwarting, Ava Soleimany, and Daniela Rus.
\newblock Deep evidential regression.
\newblock {\em NeurIPS}, 33:14927--14937, 2020.

\bibitem{dusenberry_efficient_2020}
Michael Dusenberry, Ghassen Jerfel, Yeming Wen, Yian Ma, Jasper Snoek,
  Katherine Heller, Balaji Lakshminarayanan, and Dustin Tran.
\newblock Efficient and scalable bayesian neural nets with rank-1 factors.
\newblock In {\em Proc. ICML}, pages 2782--2792, 2020.

\bibitem{fort2019deep}
Stanislav Fort, Huiyi Hu, and Balaji Lakshminarayanan.
\newblock Deep ensembles: A loss landscape perspective.
\newblock {\em arXiv:1912.02757}, 2019.

\bibitem{frankle_lottery_2019}
Jonathan Frankle and Michael Carbin.
\newblock The lottery ticket hypothesis: Finding sparse, trainable neural
  networks.
\newblock {\em arXiv:1803.03635}, 2018.

\bibitem{gal_dropout_2015}
Yarin Gal and Zoubin Ghahramani.
\newblock Dropout as a bayesian approximation: Representing model uncertainty
  in deep learning.
\newblock In {\em international conference on machine learning}, volume~48,
  pages 1050--1059. PMLR, 2016.

\bibitem{gawlikowski2021survey}
Jakob Gawlikowski, Cedrique Rovile~Njieutcheu Tassi, Mohsin Ali, Jongseok Lee,
  Matthias Humt, Jianxiang Feng, Anna Kruspe, Rudolph Triebel, Peter Jung,
  Ribana Roscher, et~al.
\newblock A survey of uncertainty in deep neural networks.
\newblock {\em arXiv:2107.03342}, 2021.

\bibitem{goodfellow_explaining_2015}
Ian~J Goodfellow, Jonathon Shlens, and Christian Szegedy.
\newblock Explaining and harnessing adversarial examples.
\newblock {\em arXiv:1412.6572}, 2014.

\bibitem{havasi_training_2021}
Marton Havasi, Rodolphe Jenatton, Stanislav Fort, Jeremiah~Zhe Liu, Jasper
  Snoek, Balaji Lakshminarayanan, Andrew~M Dai, and Dustin Tran.
\newblock Training independent subnetworks for robust prediction.
\newblock {\em arXiv:2010.06610}, 2020.

\bibitem{huang_apolloscape_2018}
Xinyu Huang, Xinjing Cheng, Qichuan Geng, Binbin Cao, Dingfu Zhou, Peng Wang,
  Yuanqing Lin, and Ruigang Yang.
\newblock The apolloscape dataset for autonomous driving.
\newblock In {\em Proc. CVPRW}, pages 1067--1073, 2018.

\bibitem{kendall2017uncertainties}
Alex Kendall and Yarin Gal.
\newblock What uncertainties do we need in bayesian deep learning for computer
  vision?
\newblock {\em NeurIPS}, 30:5580–5590, 2017.

\bibitem{kotz_laplace_2001}
Samuel Kotz, Tomasz Kozubowski, and Krzysztof Podg{\'o}rski.
\newblock {\em The Laplace distribution and generalizations: a revisit with
  applications to communications, economics, engineering, and finance}.
\newblock Number 183. Springer Science \& Business Media, 2001.

\bibitem{kuleshov_accurate_2018}
Volodymyr Kuleshov, Nathan Fenner, and Stefano Ermon.
\newblock Accurate uncertainties for deep learning using calibrated regression.
\newblock In {\em Proc. ICML}, pages 2796--2804. PMLR, 2018.

\bibitem{lakshminarayanan2017simple}
Balaji Lakshminarayanan, Alexander Pritzel, and Charles Blundell.
\newblock Simple and scalable predictive uncertainty estimation using deep
  ensembles.
\newblock {\em NeurIPS}, 30:6405–6416, 2017.

\bibitem{loshchilov_decoupled_2019}
Ilya Loshchilov and Frank Hutter.
\newblock Decoupled weight decay regularization.
\newblock In {\em Proc. ICLR}, 2017.

\bibitem{meinert_unreasonable_2022}
Nis Meinert, Jakob Gawlikowski, and Alexander Lavin.
\newblock The unreasonable effectiveness of deep evidential regression.
\newblock In {\em Proc. AAAI}, volume~37, pages 9134--9142, 2023.

\bibitem{molchanov_pruning_2017}
Pavlo Molchanov, Stephen Tyree, Tero Karras, Timo Aila, and Jan Kautz.
\newblock Pruning convolutional neural networks for resource efficient
  inference.
\newblock {\em arXiv:1611.06440}, 2016.

\bibitem{nair_maximum_2022}
Deebul~S Nair, Nico Hochgeschwender, and Miguel~A Olivares-Mendez.
\newblock Maximum likelihood uncertainty estimation: Robustness to outliers.
\newblock {\em arXiv:2202.03870}, 2022.

\bibitem{nix1994estimating}
D.A. Nix and A.S. Weigend.
\newblock Estimating the mean and variance of the target probability
  distribution.
\newblock In {\em Proc. ICNN}, volume~1, pages 55--60, 1994.

\bibitem{Paszke_PyTorch_An_Imperative_2019}
Adam Paszke, Sam Gross, Francisco Massa, Adam Lerer, James Bradbury, Gregory
  Chanan, Trevor Killeen, Zeming Lin, Natalia Gimelshein, Luca Antiga, Alban
  Desmaison, Andreas Kopf, Edward Yang, Zachary DeVito, Martin Raison, Alykhan
  Tejani, Sasank Chilamkurthy, Benoit Steiner, Lu Fang, Junjie Bai, and Soumith
  Chintala.
\newblock {PyTorch: An Imperative Style, High-Performance Deep Learning
  Library}.
\newblock In {\em Proc. NeurIPS}, volume~32, pages 8024--8035. Curran
  Associates, Inc., 2019.

\bibitem{ronneberger_u-net_2015}
Olaf Ronneberger, Philipp Fischer, and Thomas Brox.
\newblock U-net: Convolutional networks for biomedical image segmentation.
\newblock In {\em MICCAI}, volume~18, pages 234--241. Springer, 2015.

\bibitem{rosberg_globally_2023}
Thomas Ro{\ss}berg and Michael Schmitt.
\newblock A globally applicable method for ndvi estimation from sentinel-1 sar
  backscatter using a deep neural network and the sen12tp dataset.
\newblock {\em PFG}, 91:171--188, 2023.

\bibitem{silberman_indoor_2012}
Nathan Silberman, Derek Hoiem, Pushmeet Kohli, and Rob Fergus.
\newblock Indoor segmentation and support inference from rgbd images.
\newblock In {\em ECCV}, volume~12, pages 746--760. Springer, 2012.

\bibitem{soleimany2021evidential}
Ava~P Soleimany, Alexander Amini, Samuel Goldman, Daniela Rus, Sangeeta Bhatia,
  and Connor Coley.
\newblock Evidential deep learning for guided molecular property prediction and
  discovery.
\newblock {\em ACS Central Science}, 7:1356--1367, 2021.

\bibitem{srivastava2014dropout}
Nitish Srivastava, Geoffrey Hinton, Alex Krizhevsky, Ilya Sutskever, and Ruslan
  Salakhutdinov.
\newblock Dropout: A simple way to prevent neural networks from overfitting.
\newblock {\em JMLR}, 15:1929--1958, 2014.

\bibitem{welling2011bayesian}
Max Welling and Yee~W Teh.
\newblock Bayesian learning via stochastic gradient langevin dynamics.
\newblock In {\em Proc. ICML}, pages 681--688, 2011.

\bibitem{wen_batchensemble_2020}
Yeming Wen, Dustin Tran, and Jimmy Ba.
\newblock Batchensemble: an alternative approach to efficient ensemble and
  lifelong learning.
\newblock {\em arXiv:2002.06715}, 2020.

\end{thebibliography}
}

\end{document}